\documentclass{article}\usepackage[]{graphicx}\usepackage[]{xcolor}
\makeatletter
\def\maxwidth{ %
  \ifdim\Gin@nat@width>\linewidth
    \linewidth
  \else
    \Gin@nat@width
  \fi
}
\makeatother

\definecolor{fgcolor}{rgb}{0.345, 0.345, 0.345}

\usepackage{framed}
\makeatletter
 {\par\unskip\endMakeFramed%
 \at@end@of@kframe}
\makeatother

\definecolor{shadecolor}{rgb}{.97, .97, .97}
\definecolor{messagecolor}{rgb}{0, 0, 0}
\definecolor{warningcolor}{rgb}{1, 0, 1}
\definecolor{errorcolor}{rgb}{1, 0, 0}
\newenvironment{knitrout}{}{} % an empty environment to be redefined in TeX

\usepackage{alltt} 
\usepackage{graphicx}
\usepackage{natbib} 
\usepackage{amsmath} %align
\usepackage{bm} %bf
\usepackage{comment} %comment
\usepackage[]{todonotes}%\todo
\usepackage{url}
\usepackage{lineno}
\usepackage{authblk}
\usepackage{booktabs}
\usepackage{hyperref}

\def\min{\mathop{\rm min}}

\usepackage{amssymb} %mathbb
\usepackage[symbol]{footmisc} %%footnote
\usepackage[title]{appendix}%"Appendix" comes out in appendix

\usepackage[boxed]{algorithm2e}
\newcommand{\ma}[1]{\ensuremath{\mathbf{#1}}}

\newcommand{\tr}{^{\sf T}}

\SetKwComment{Comment}{/*~}{~*/}

\usepackage{listings}
\usepackage{xcolor}

\definecolor{codegreen}{rgb}{0,0.6,0}
\definecolor{codegray}{rgb}{0.5,0.5,0.5}
\definecolor{codepurple}{rgb}{0.58,0,0.82}
\definecolor{backcolour}{rgb}{0.9215686,0.9215686,0.9215686}

\lstdefinestyle{mystyle}{
    backgroundcolor=\color{backcolour},   
    commentstyle=\color{codegreen},
    keywordstyle=\color{magenta},
    numberstyle=\tiny\color{codegray},
    stringstyle=\color{codepurple},
    basicstyle=\ttfamily\footnotesize,
    breakatwhitespace=false,         
    breaklines=true,                 
    captionpos=b,                    
    keepspaces=true,                 
    numbers=left,                    
    numbersep=5pt,                  
    showspaces=false,                
    showstringspaces=false,
    showtabs=false,                  
    tabsize=2
}

\IfFileExists{upquote.sty}{\usepackage{upquote}}{}
\begin{document}

\title{Interpretable Kernels}

\author[1]{Patrick J.F. Groenen}
\author[2]{Michael Greenacre}
\affil[1]{Econometric Institute, Erasmus University Rotterdam,
              The Netherlands}
\affil[2]{Universitat Pompeu Fabra, Barcelona, Spain}

\date{\today}

\maketitle

\begin{abstract}
The use of kernels for nonlinear prediction is widespread in machine learning. 
They have been popularized in support vector machines and used in kernel ridge regression, amongst others. 
Kernel methods share three aspects. 
First, instead of the original 
matrix of predictor variables or features, each observation is mapped into an enlarged feature space. 
Second, a ridge penalty term is used to shrink the coefficients on the features in the enlarged feature space. 
Third, the solution is not obtained in this enlarged feature space, but through solving a dual problem in the observation space.  
A major drawback in the present use of kernels is that the interpretation in terms of the original features is lost. 
In this paper, we argue that in the case of a wide 
matrix of features, 
where there are more features than observations, the kernel solution can be re-expressed in terms of a linear combination of the original matrix of features and a ridge penalty that involves a special metric. 
Consequently, the exact same predicted values can be obtained as a weighted linear combination of the features in the usual manner and thus can be interpreted. 
In the case where the number of features is less than the number of observations, we discuss a least-squares approximation of the kernel matrix that still allows the interpretation in terms of a linear combination. 
It is shown that these results hold for any function of a linear combination that minimizes the coefficients and has a ridge penalty on these coefficients, such as in kernel logistic regression and kernel Poisson regression. 
This work makes a contribution to interpretable artificial intelligence.
\end{abstract}

\section{Introduction}
\label{intro}
The last two decades have shown a dramatic improvement of prediction accuracy by artificial intelligence methods through machine learning. The availability of big data combined with highly nonlinear models has led to improved prediction capabilities. However, in many applications, such nonlinear predictions cannot be used due to regulatory requirements. For example, a bank cannot use directly the results from a deep neural net ``black box" algorithm  for credit scoring because the regulator requires the bank to be able to explain their credit scoring rules. Therefore, explainable artificial intelligence (abbreviated as XAI) has received much attention. 
\cite{ali2023XAI} and \cite{Dwivedi2023XAI} give overviews of XAI, classifying the area into different types, where the present contribution would be included in the category of ``model explainability".
Furthermore, our approach falls in what \cite{rudin2019blackbox} calls models that are ``inherently interpretable" rather than explaining black box models. 

This paper focuses on kernel methods that are used for nonlinear predictions, where our main contribution lies in re-expressing the kernel penalty term such that a linear interpretation in the features becomes possible. 
The use of kernels has become a popular and attractive tool to allow for nonlinear predictions. It has gained popularity in machine learning through the method of support vector machines and can be applied in many other techniques that use linear prediction. We refer to \cite{Hastie2007}, Section 5.8, who give a simplified introduction to kernel methods based on the explanation in \cite{wahba1990spline}. We provide below a linear algebra interpretation.

The main idea is to map each row $\ma{x}_i\tr$ of the $n \times p$ matrix $\ma{X}$ of features into the vector $\bm{\phi}_i\tr$ of an enlarged feature space represented by the $n\times r$ matrix $\bm{\Phi}$, where $r\gg p$. 
A well-known example is the so-called second degree nonhomogeneous polynomial kernel that is equivalent to choosing  $\boldsymbol{\phi}_i\tr$ as the vector of all main and two-variable interaction effects, which is a special case of the features in polynomial regression. 
Then, the linear prediction is done in the enlarged feature space of the $\bm{\phi}_i\tr$ by $\bm{\phi}_i\tr\bm{\beta}$, which is equivalent to a nonlinear prediction in the original space of the $\ma{x}_i\tr$.  
As the dimensionality $r$ of the space spanned by $\boldsymbol{\phi}_i\tr$ increases towards or beyond the number of observations (rows) $n$, then some form of shrinkage is needed to avoid overfitting. 

For very large $r$, the estimation of explicit coefficients $\bm{\beta}$ of the linear combination $\bm{\eta} = \bm{\Phi}\bm{\beta}$ becomes infeasible. 
This situation is one where the so-called kernel trick can be used: that is, instead of computing the solution for $\bm{\beta}$, the optimization can be done directly over the predictions in $\bm{\eta}$, making use of the $n \times n$ kernel matrix $\ma{K} = \bm{\Phi\Phi}\tr$. 
Mappings from $\ma{X}$ to $\bm{\Phi}$ that satisfy the so-called Mercer conditions \citep{mercer1970mercer, campbell2002kernel} have very simple expressions for $k_{ij}(\ma{x}_i, \ma{x}_j)$ as a function of $\ma{x}_i$ and $\ma{x}_j$ only (which we will also denote by $k_{ij}$ to simplify notation).
For example, the radial basis function (RBF) kernel (or Gaussian kernel) has $k_{ij} = \exp({-s \|\ma{x}_i - \ma{x}_j\|^2})$ (with $s>0$ a scaling hyper-parameter) and the nonhomogeneous polynomial kernel of degree $d$ has $k_{ij} = (1 + \ma{x}_i\tr \ma{x}_j)^d$. 

One of the main disadvantages of the use of kernels is that only the resulting predictions $\bm{\eta} = \bm{\Phi}\bm{\beta}$ are found, without knowing $\bm{\beta}$.  
Thus, for many kernels no model interpretation is given in terms of the derived features in $\bm{\Phi}$ or in the original features in $\ma{X}$. 
This paper offers a general solution to this problem. 
Here, it is shown that if $p \geq n$ the kernel solution can also be expressed in terms of the original features, that is, $\bm{\eta} = \bm{\Phi}\bm{\beta} = \ma{X}\bm{\gamma}$, so that the coefficients in $\bm{\gamma}$ can be interpreted similarly as in regression. In this case, both linear combinations and both kernel penalties are exactly the same.  
For $p < n$, we propose to use a least-squares approximation of the kernel penalty that preserves the prediction by $\bm{\eta} = \ma{X}\bm{\gamma}$ and thus the linear interpretation through the coefficients in $\bm{\gamma}$. A simple diagnostic is proposed, the kernel accounted for (KAF), that gives the proportion of the kernel penalty that is accounted for by the approximation.

Next to the explainability of the kernel, a second important conclusion of this paper is that the nonlinear prediction through kernels critically depends on the kernel penalty and is equivalent to a linear prediction with a quadratic penalty on the coefficients, with contributions to the penalty not only on the size of the individual coefficients (as for the ridge penalty) but also on their combinations. 

A third contribution is the realization that the linear interpretation proposed in this paper can be applied to any prediction model that uses a kernel in combination with a quadratic kernel penalty. Examples are support vector machines, kernel ridge regression, kernel logistic regression, kernel Poisson regression, amongst others. With this generality, we contribute to explainable AI.

The present paper can be seen as a contribution following the call of \cite{crawford2018kernel} who provided general approximation for kernels and what they call effect size analogs through projections in Bayesian Approximate Kernel Regression. They write: ``It should be clear that a variety of projection procedures can be specified corresponding to various priors and loss functions, and a systematic study elucidating which projections are efficient and robust is of great interest.'' We follow a linear algebra derivation leading to coefficients that are exactly the same in cases of a wide matrix of predictors, and approximate otherwise.

Section~\ref{sect:approx_kernel} starts with the general theory of a kernel penalty and its approximation, showing how coefficients for the original variables can be obtained. 
Section~\ref{sect:Computation} deals with computational aspects, where it is shown how solutions to the kernel approach can be obtained using standard software. 
%Section 4 deals with the application to ridge regression and its kernel version using the ridge penalty, that is the sum of the squared predictor coefficients.
%Section 4 extends the approach to kernel versions of generalized linear models. 
%and how approximate standard errors of the interpretable coefficients can be computed.
Section~\ref{sect:application} gives two applications, one in a linear regression context and the other in logistic regression.
Section~\ref{sect:conclusions} concludes with a summary and discussion.

\section{Approximate kernel ridge regression and estimation of interpretable regression coefficients}
\label{sect:approx_kernel}

%--------------------------------------------------------
\subsection{Kernel ridge regression and the dual problem}
\label{sec:kernelridgedual}
In general, kernel methods can be written as the minimization of the sum of a loss function, $f(\bm{\eta})$, where $\bm{\eta} = \bm{\Phi}\bm{\beta}$, and a ridge penalty on the coefficients $\bm{\beta}$ ($r\times 1$), that is,
\begin{eqnarray}
   \min_{\bm{\beta}} \  f(\bm{\eta}) + \lambda \|\bm{\beta}\|^2, \label{for:def_g}
\end{eqnarray}
where $\|\bm{\beta}\|^2 = \bm{\beta}\tr\bm{\beta}$ and $\lambda>0$ is a fixed penalty strength parameter that can be determined through $k$-fold cross-validation. 
The simplest example is ridge regression with $f(\bm{\eta}) = \|\ma{y} - \bm{\eta}\|^2$, where $\ma{y}$ is an observed $n\times 1$ vector of a response variable, $\bm{\Phi}$ is chosen as the observed predictor matrix $\ma{X}$, and so $\bm{\eta} = \ma{Xb}$.
This scheme includes generalized linear models where the function $f$ is the deviance and there is a link function $g(\bm{\eta})$ between the linear model and the response mean.

Our approach starts by passing from the original minimization problem in terms of coefficients $\bm{\beta}$ in the linear combinations $\bm{\eta} = \bm{\Phi}\bm{\beta}$ of the high-dimensional predictors, to the dual problem of estimating  $\bm{\eta}$ itself where the penalty is imposed on $\bm{\eta}$.

Hence, $\|\bm{\beta}\|^2$ is rewritten as a quadratic function of $\bm{\eta}$, by solving for $\bm{\beta}$ as follows: 
\begin{eqnarray}
    \bm{\eta} &=& \bm{\Phi}\bm{\beta} \nonumber \\
    \bm{\Phi}\tr\bm{\eta} &=& \bm{\Phi}\tr\bm{\Phi}\bm{\beta} \nonumber \\
    (\bm{\Phi}\tr\bm{\Phi})^-\bm{\Phi}\tr\bm{\eta} &=& \bm{\beta} ,
    \label{for:b}
\end{eqnarray}
where $(\ )^-$ refers to the regular matrix inverse, if it exists, otherwise the Moore-Penrose generalized inverse.
Then, \renewcommand{\thefootnote}{\arabic{footnote}}
\begin{eqnarray}
    \|\bm{\beta}\|^2&=& \bm{\beta}\tr\bm{\beta} \nonumber \\
    &=& \bm{\eta}\tr \bm{\Phi}(\bm{\Phi}\tr\bm{\Phi})^- 
    (\bm{\Phi}\tr\bm{\Phi})^-\bm{\Phi}\tr\bm{\eta} \nonumber \\
    &=& \bm{\eta}\tr (\bm{\Phi}\bm{\Phi}\tr)^{-}\bm{\eta} \nonumber \\
    &=& \bm{\eta}\tr\ma{K}^{-}\bm{\eta} .
    \label{for:Penalty}
\end{eqnarray} 
Notice that the step from the second to the third lines in (\ref{for:Penalty}) above can be deduced using the singular value decomposition (SVD)\footnote[1]{
    {\rm SVD:\ \ } $\bm{\Phi} = \ma{U}\bm{\Sigma}\ma{V}\tr,  \quad \ma{U} (n\times k), \ma{V} (r\times k), \bm{\Sigma} (k\times k)$,  where $\bm{\Phi}$ is of rank $k$, $\ma{U}\tr\ma{U} = \ma{V}\tr\ma{V} = \ma{I}$ and $\bm{\Sigma}$ is a diagonal matrix of positive singular values. 
 Then $(\bm{\Phi}\tr\bm{\Phi})^-  = (\ma{V}\bm{\Sigma} \ma{U}\tr \ma{U}\bm{\Sigma} \ma{V}\tr)^- 
= (\ma{V}\bm{\Sigma}^2 \ma{V}\tr)^- = \ma{V}\bm{\Sigma}^{-2} \bf{V}\tr$, where $\bm{\Sigma}^{-2}$  
 denotes a diagonal matrix of the inverses of the squares of the singular values. Similarly, $(\bm{\Phi}\bm{\Phi}\tr)^-  = \ma{U}\bm{\Sigma}^{-2} \ma{U}\tr$. The result $\bm{\Phi}(\bm{\Phi}\tr\bm{\Phi})^- 
    (\bm{\Phi}\tr\bm{\Phi})^-\bm{\Phi}\tr = (\bm{\Phi}\bm{\Phi}\tr)^-$ follows.} of $\bm{\Phi}$.
The dual problem is thus, in terms of $\bm{\eta}$, the minimization of
\begin{eqnarray}
     f(\bm{\eta}) + \lambda \bm{\eta}\tr\ma{K}^{-}\bm{\eta} \label{for:def_g_eta}
\end{eqnarray}
over $\bm{\eta}$.

We now show how to obtain estimates of interpretable coefficients $\bm{\gamma}$ ($p \times 1$) in linear combinations $\ma{X}\bm{\gamma}$, of the original predictors, which are equal to or approximate those in $\bm{\Phi}\bm{\beta}$, depending on whether $p \geq n$ or $p < n$ respectively.
This is possible thanks to approximating the kernel, and the estimation will be as successful as the closeness of the approximation.

\subsection{Approximating the kernel} 
\label{sec:approximatekernel}

The approximation to the kernel matrix $\ma{K}$ amounts to a double projection onto the rows and columns of $\ma{X}$.
There are two slightly different outcomes depending on whether $p < n$ or $p \geq n$.
Let us suppose that $\bm{\Phi}$ can be approximated by $\ma{XB}$, where $\ma B$ %($p\times r$) 
is a matrix of regression coefficients.
Since $\ma K = \bm{\Phi}\bm{\Phi}\tr$ is then approximated by $\ma{XBB}\tr \ma{X}\tr$, we are rather concerned with estimating $\ma{BB}\tr$, which can be denoted by the $p\times p$ matrix $\ma{A} = \ma{BB}\tr$.
The least-squares approximation of the kernel matrix $\ma K$ by $\ma{XA}\ma{X}\tr$ implies the following minimization objective:
\begin{equation}
  \min_{\ma{A}}\| {\ma K} - {\ma X}{\ma A}{\ma X}\tr \|^2 . \label{for:ApproxK}
\end{equation}

To minimize (\ref{for:ApproxK}) we take its partial derivatives with respect to $\ma{A}$ and equate that to zero, in order to obtain an estimate $\widehat{\ma A}$ of $\ma{A}$, that is,
\begin{eqnarray*}
     2\ma{X}\tr\ma{X} \widehat{\ma{A}} \ma{X}\tr\ma{X} - 2 \ma{X}\tr \ma{KX} &=& \ma{0} 
\end{eqnarray*}
and solve for $\widehat{\ma{A}}$
\begin{eqnarray}
  \widehat{\ma{A}} &=& ({\ma X}\tr{\ma X})^{-} {\ma X}\tr  {\ma K}\ma{X} ({\ma X}\tr{\ma X})^{-}
  \label{for:BBt}
\end{eqnarray}
so that 
\begin{eqnarray*}
    \widehat{\ma{K}}&=&\ma{X}\widehat{\ma{A}} \ma{X}\tr 
    = {\ma X}({\ma X}\tr{\ma X})^{-} {\ma X}\tr  {\ma K}\ma{X} ({\ma X}\tr{\ma X})^{-}{\ma X}\tr
\end{eqnarray*}
is the approximate kernel matrix. Note that ${\ma X}({\ma X}\tr{\ma X})^{-} {\ma X}\tr$ is a projector matrix so that $\widehat{\ma{K}}$ can be interpreted as the projection of the row and column spaces of $\ma{K}$ onto the space spanned by $\ma{X}$.

The amount of loss incurred by the penalty approximation can be obtained by inserting $\widehat{\ma{A}}$ from (\ref{for:BBt}) into (\ref{for:ApproxK}), that is,
\begin{eqnarray}
    \| {\ma K} - \widehat{\ma{K}}\|^2&=& 
    \| {\ma K} - {\ma X}\widehat{\ma{A}}{\ma X}\tr \|^2 \nonumber \\ 
    &=&     \| {\ma K} - {\ma X}({\ma X}\tr{\ma X})^{-} {\ma X}\tr  {\ma K}\ma{X} ({\ma X}\tr{\ma X})^{-}{\ma X}\tr \|^2 \nonumber \\
    &=& \|\left({\ma I} - {\ma X}({\ma X}\tr{\ma X})^{-} {\ma X}\tr)\ma{K}({\ma I} - {\ma X}({\ma X}\tr{\ma X})^{-} {\ma X}\tr\right)\|^2 \label{for:PenApproxLoss}
\end{eqnarray}
from which it can be seen that the loss is equal to the part of $\ma{K}$ that is not in the space of $\ma{X}$ .

From (\ref{for:PenApproxLoss}), we define the \textit{kernel accounted for} (KAF) as the proportion of ${\|\ma{K}\|^2}$ in the space of 
$\ma{X}$:
\begin{eqnarray}
   \textrm{KAF} =  \frac{\| \widehat{\ma{K}}\|^2}{\|\ma{K}\|^2}\label{for:KAF} .
\end{eqnarray}
Note that if $p \geq n$ and $\ma{X}$ is of full rank then, from (\ref{for:BBt}) and again using the SVD, the approximation $\widehat{\ma{K}}$ is perfect and the objective criterion in (\ref{for:ApproxK}) is zero and $\text{KAF} = 1$; this is the exact case.
If $p < n$, then $\text{KAF}< 1$; this is the approximate case.

\subsection{Estimating the interpretable coefficients} 
\label{sec:coefficients}

Finally, we arrive at estimating $p$ interpretable coefficients on the original $p$ predictors. 
The results are different depending on whether $p\geq n$ or $p<n$.
In the former exact case, $\bm{\Phi}$ can be written as  $\ma{XB}$, where $\ma{B}$ is of full rank $n$.
Thus, the linear combination $\bm{\eta} = \bm{\Phi} \bm{\beta}$ is equal to $\ma{XB}\bm{\beta}$, which can be written as ${\ma X}\bm{\gamma}$, 
with coefficients on the original features (columns of $\ma X$) equal to $\bm{\gamma} =\ma{B}\bm{\beta}$.
Solving for $\bm{\beta}$, $\bm{\beta} = (\ma{B}\tr\ma{B})^{-} \ma{B}\tr\bm{\gamma}$ (as obtained in (2)),  
and the penalty in (\ref{for:def_g_eta}) in this exact case can be expressed in terms of $\bm{\beta}$ as
\begin{equation}
    \bm{\beta}\tr\bm{\beta}  = \bm{\gamma}\tr \ma{B} (\ma{B}\tr\ma{B})^{-}(\ma{B}\tr\ma{B})^{-}\ma{B}\tr \bm{\gamma} =\bm{\gamma}\tr (\ma{BB}\tr)^{-}\bm{\gamma}
    = \bm{\gamma}\tr \widehat{\ma{A}}^{\!-} \bm{\gamma},
    \label{exactPenalty}
\end{equation}
(see the simplification using the SVD in Footnote 1).

In the approximate case when $p<n$, $\bm{\Phi}$ is only approximated by $\ma{XB}$, and the result in (\ref{exactPenalty}) reduces to the approximate penalty 
$\bm{\beta}\tr\bm{\beta}  \approx \bm{\gamma}\tr \widehat{\ma{A}}^{\!-}\bm{\gamma}$.
%Finally, we arrive at $p$ interpretable coefficients on the original $p$ predictors. 
%Since $\bm{\Phi}$ is approximated by $\ma{XB}$, the linear combination $\bm{\eta} = \bm{\Phi} \bm{\beta}$ are approximated by $\ma{XB}\widetilde{\bm{\beta}} = {\ma X}\bm{\gamma}$, 
%where the coefficients of the original features (columns of $\ma X$) can be written as $\bm{\gamma} = \ma{B}\widetilde{\bm{\beta}}$.
%Solving $\widetilde{\bm{\beta}} = (\ma{B}\tr\ma{B})^{-} \ma{B}\tr\bm{\gamma}$ (as obtained in (2)),  
%the penalty in (\ref{for:def_g_eta}) in the exact case can thus be expressed in terms of $\tilde{\bm{\beta}}$ as
%\begin{eqnarray}
%    \bm{\beta}\tr\bm{\beta}  = \widetilde{\bm{\beta}%}\tr\widetilde{\bm{\beta}}&=& \bm{\gamma}\tr %(\ma{BB}\tr)^{-}\bm{\gamma}\nonumber \\
%    &=& \bm{\gamma}\tr \widehat{\ma{A}}^{\!-}\bm{\gamma},
%\end{eqnarray}
%\noindent
%or in the approximate case the penalty itself is approximated as
%\begin{eqnarray}
%    \bm{\beta}\tr\bm{\beta}  \approx \widetilde{\bm{\beta}}\tr\widetilde{\bm{\beta}} &=& \bm{\gamma}\tr \widehat{\ma{A}}^{\!-}\bm{\gamma}.
%\end{eqnarray}
In the exact case, we have the equality of the kernel solution $\bm{\eta}$, which is numerically the same as the solution of $\ma{X}\bm{\gamma}$ with penalty $\lambda \bm{\gamma}\tr\widehat{\ma{A}}^{-}\bm{\gamma}$, that is,
\begin{eqnarray}
      f(\bm{\eta}) + \lambda \bm{\eta}\tr\ma{K}^{-1}\bm{\eta} &=& f(\ma{X}\bm{\gamma}) + \lambda \bm{\gamma}\tr\widehat{\ma{A}}^{-}\bm{\gamma}.
 \label{for:equivalence}
\end{eqnarray}
Therefore, in this exact case, it is possible to obtain the same kernel solution by a traditional linear combination where the coefficients $\bm{\gamma}$ can be interpreted as in traditional regression techniques. 
In the approximate case, the kernel solution is approximate, with its quality measured by the KAF.

%For purposes of out-of-sample prediction for $n_{\text{test}}$ observations in the $n_{\text{test}} \times n$ matrix $\ma{X}_{\text{test}}$ of test data, the linear combinations can be easily obtained through $\bm{\eta}_{\text{test}} = \ma{X}_{\text{test}} \bm{\gamma}$.

%-------------------------------------------------
\subsection{Dealing with the regression intercept}
\label{sec:intercept}
The derivations above assume that if there is an intercept, then the intercept is modelled through a column of ones in $\ma{X}$ and the intercept itself is penalized in the ridge penalty. In many cases, one may want to avoid the regularization of the intercept, or simply not estimate the intercept at all. As many kernels implicitly estimate the intercept, some extra steps are required to avoid estimation of the intercept in the linear combination. Let $\ma{J} = \ma{I}- n^{-1}\ma{11}\tr$ be the $n \times n$ centring matrix. Then, $\ma{JX}$ is the column centred version of $\ma{X}$ and, similarly, $\ma{J}\bm{\Phi}$ is the column centred high-dimensional space with $\ma{J}\bm{\Phi}\bm{\Phi}\tr\ma{J} = \ma{JKJ} = \ma{K}_{\text{c}}$, the double centred version of $\ma{K}$. We use the subscript c to denote the (double) centred versions of a matrix or vector. The double centred approximation $\ma{K}_{\text{c}}$ through the minimization of $\|\ma{K}_{\text{c}} - \ma{X}_{\text{c}} \ma{A}\ma{X}_{\text{c}}\tr \|^2$ over $\ma{A}$ with $\ma{X}_{\text{c}} = \ma{JX}$ is achieved by 
\begin{eqnarray}
  \widehat{\ma{A}}_{\text{c}} &=& (\ma{X}_{\text{c}}\tr\ma{X}_{\text{c}})^{-} \ma{X}_{\text{c}}\tr  \ma{K}_{\text{c}}\ma{X}_{\text{c}} (\ma{X}_{\text{c}}\tr\ma{X}_{\text{c}})^{-}.
  \label{for:BBtcentered}
\end{eqnarray}
The equivalent of (\ref{for:equivalence}) with an unpenalized intercept $\alpha$ and centred $\bm{\eta}_{\text{c}} = \ma{J}\bm{\eta}$ then becomes
\begin{eqnarray}
      f(\alpha \ma{1} + \bm{\eta}_{\text{c}}) + \lambda \bm{\eta}_{\text{c}}\tr\ma{K}_{\text{c}}^-\bm{\eta}_{\text{c}} &=& f(\alpha \ma{1} + \ma{X}_{\text{c}}\bm{\gamma}) + \lambda \bm{\gamma}\tr\widehat{\ma{A}}_{\text{c}}^{-}\bm{\gamma}.
 \label{for:equivalenceCentered}
\end{eqnarray}

%--------------------------------------------------------
\subsection{Re-expresssing the kernel penalty as a regular ridge penalty}
\label{sect:reexpressKernelAsRegularRidge}
As for the exact case, it is also possible to re-express the kernel approach as a linear combination $\ma{Z}\bm{\delta}$ with ridge penalty term $\lambda \bm{\delta}\tr \bm{\delta}$. 
The advantage of this re-expression is that the kernel solution results can be computed using standard software (e.g., the \texttt{glmnet} package \citep{Friedman:10, Tay:23} in \textsf{R} \citep{R:22}) which is based on a linear combination using features in $\ma{Z}$, coefficients $\bm{\delta}$, and a ridge penalty. Below, the pre- and post-processing calculations are given. 

Let $\ma{K} = \ma{QDQ}\tr$ be the eigendecomposition of $\ma{K}$, with $\ma{D}$ the diagonal matrix of positive eigenvalues so that $\ma{K}$ is assumed to be positive definite and $\ma{Q}$ the matrix of corresponding eigenvectors with $\ma{Q}\tr\ma{Q} = \ma{I}$, and also $\ma{Q}\ma{Q}\tr = \ma{I}$ since $\ma{Q}$ is square. Then,
\begin{eqnarray}
    \lambda \bm{\eta}\tr\ma{K}^{-1}\bm{\eta} = \lambda (\bm{\eta}\tr\ma{QD}^{-1/2})(\ma{D}^{-1/2}\ma{Q}\tr\bm{\eta}) = \lambda\bm{\delta}\tr \bm{\delta}
\end{eqnarray}
using 
\begin{eqnarray*}
    \bm{\delta}&=&\ma{D}^{-1/2}\ma{Q}\tr\bm{\eta} 
\end{eqnarray*}
so that
\begin{eqnarray*}
  \bm{\eta}&=& \ma{Q}\ma{D}^{1/2}\bm{\delta} = \ma{Z}\bm{\delta}.
\end{eqnarray*}
Note that $\ma{Z} = \ma{QD}^{1/2}$ also implies that $\ma{Z} = (\ma{QDQ})\tr\ma{QD}^{-1/2} = \ma{KQD}^{-1/2}$.

In the case of approximate kernels, use the approximate kernel $\widehat{\ma{K}} = {\ma X}\widehat{\ma{A}}{\ma X}\tr$ instead of $\ma{K}$. Then
\begin{eqnarray*}
    \bm{\eta} &=& \ma{Z}\bm{\delta} \\
    &=&  \widehat{\ma{K}}\ma{QD}^{-1/2}\bm{\delta} \\
    &=&  (\ma{X} \widehat{\ma{A}}\ma{X}\tr) \ma{QD}^{-1/2}\bm{\delta} \\
    &=&  \ma{X} (\widehat{\ma{A}}\ma{X}\tr \ma{QD}^{-1/2}\bm{\delta}) \\
    &=&  \ma{X} \bm{\gamma}
\end{eqnarray*}
so that $\bm{\gamma} = \widehat{\ma{A}}\ma{X}\tr \ma{QD}^{-1/2}\bm{\delta}$.
In this way, we obtain coefficient estimates that can be interpreted as quantifying effect sizes of the predictor variables on the response.

\subsection{Test set predictions}

Often there is the need to provide predicted values for unseen data in the test set $\ma{X}_\text{test}$ with $n_{\text{test}}$ rows. With kernels, the prediction becomes
\begin{eqnarray*}
    \bm{\eta}_{\text{test}} &=& \bm{\Phi}_{\text{test}}\bm{\beta} \nonumber \\
     &=& \bm{\Phi}_{\text{test}}(\bm{\Phi}\tr\bm{\Phi})^-\bm{\Phi}\tr\bm{\eta} 
\end{eqnarray*}
Using the SVD of $\bm{\Phi}$ from Footnote 1, $\bm{\eta}_{\text{test}}$ becomes
\begin{eqnarray*}
    \bm{\eta}_{\text{test}} &=&  \bm{\Phi}_{\text{test}} (\ma{V} \bm{\Sigma}^{-2} \ma{V}\tr)(\ma{V} \bm{\Sigma} \ma{U}\tr) \bm{\eta} \nonumber \\
    &=& \bm{\Phi}_{\text{test}} \ma{V} \bm{\Sigma}^{-1}  \ma{U}\tr \bm{\eta} \nonumber \\
    &=& \bm{\Phi}_{\text{test}} (\ma{V} \bm{\Sigma}  \ma{U}\tr) (\ma{U} \bm{\Sigma}^{-2}  \ma{U}\tr) \bm{\eta} \nonumber \\
    &=& \bm{\Phi}_{\text{test}} \bm{\Phi}\tr (\bm{\Phi}\bm{\Phi}\tr)^- \bm{\eta} \nonumber \\
    &=& \ma{K}_{\text{test}} \ma{K}^{-1} \bm{\eta}, \label{for:TestPredClas}
\end{eqnarray*}
where $\ma{K}_{\text{test}} = \bm{\Phi}_{\text{test}} \bm{\Phi}\tr$ is the between-training-test-block kernel matrix. $\ma{K}_{\text{test}}= \bm{\Phi}_{\text{test}}\bm{\Phi}\tr$ and has elements  $k_{ij}(\ma{x}_{\text{test},i}, \ma{x}_j)$ where $\ma{x}_{\text{test},i}\tr$ denotes row $i$ of $\ma{X}_{\text{test}}$. The prediction using kernels can be obtained by using $\ma{Z}_{\text{test}} = \ma{K}_{\text{test}}\ma{QD}^{-1/2}$ and the linear combination for the test data becomes 
\begin{eqnarray*}
    \bm{\eta}_{\text{test}}&=&\ma{Z}_{\text{test}} \bm{\delta} \\
    &=& \ma{K}_{\text{test}} \ma{K}^{-1}\ma{X}\bm{\gamma}.
\end{eqnarray*}
Note that this prediction is \textit{not} a linear combination of $\ma{X}_{\text{test}}$ as it maps $\ma{X}_{\text{test}}$ first to the high dimensional space of $\bm{\Phi}_{\text{test}}$.

Consider the alternative prediction for the test data using the approximate kernel approach as a linear combination from the training set coefficients $\bm{\gamma}$, that is,
\begin{eqnarray*}
    \bm{\eta}_{\text{appr.test}}&=&\ma{X}_{\text{test}} \bm{\gamma}.
\end{eqnarray*}
For the test data, this prediction comes at the cost that the reconstruction of $\ma{K}_{\text{test}}$ is not optimal, not even in the case of a KAF of 1 for the training data. This can be seen as follows:   
\begin{eqnarray*}
    \bm{\eta}_{\text{appr.test}}&=&\ma{X}_{\text{test}} \bm{\gamma}\\
    &=&\ma{X}_{\text{test}} \widehat{\ma{A}}\ma{X}\tr \ma{QD}^{-1/2}\bm{\delta}
\\
    &=&\ma{X}_{\text{test}} \widehat{\ma{A}}\ma{X}\tr (\ma{X}\widehat{\ma{A}} \ma{X}\tr)^{-1}\bm{\eta}\\
    &=&[\ma{X}_{\text{test}} ({\ma X}\tr{\ma X})^{-} {\ma X}\tr  {\ma K}\ma{X} ({\ma X}\tr{\ma X})^{-} \ma{X}\tr]\widehat{\ma{K}}^{-}\bm{\eta}.
%\widehat{\ma{A}} &=& ({\ma X}\tr{\ma X})^{-} {\ma X}\tr  {\ma K}\ma{X} ({\ma X}\tr{\ma X})^{-}
\end{eqnarray*}
This means that $\ma{K}_{\text{test}}$ is approximated by
\begin{eqnarray*}
    \widehat{\ma{K}}_{\text{test}} = \ma{X}_{\text{test}} ({\ma X}\tr{\ma X})^{-} {\ma X}\tr  {\ma K}\ma{X} ({\ma X}\tr{\ma X})^{-} \ma{X}\tr,
\end{eqnarray*}
the approximate test kernel matrix $\widehat{\ma{K}}_{\text{test}}$.
Note that this approximation is based on $\ma{K}$ only and does not make use of $\ma{K}_{\text{test}}$.
% 05-08-2025 PG: I don't think we should keep KAF_test as there is no guarantee that it is between 0 and 1.
% As for KAF, the fit of this approximation can be expressed as
%   $ \text{KAF}_{\text{test}} = \|\widehat{\ma{K}}_{\text{test}}\|^2 / \|\ma{K}_{\text{test}} \|^2$. 

\section{Computational aspects}
\label{sect:Computation}
Using the results of the previous section, a kernel solution using both exact and approximated kernels can be obtained by any standard software such as \texttt{glmnet},
%in R \citep{Friedman:10,Tay:23}
using the following calculations.
\begin{enumerate}
    \item Compute the kernel matrix $\ma{K}$, using the chosen kernel, for example the RBF kernel.
    \item Compute the eigendecomposition $\ma{K} = \ma{QDQ}\tr$.
    \item Compute $\ma{Z} = \ma{Q}\ma{D}^{1/2}$
    \item Estimate $\bm{\delta}$ through standard software that has a ridge penalty using $\ma{Z}$ as features.
    \item Compute the linear combination as $\bm{\eta} = \ma{Z}\bm{\delta}$.
    \item Compute the feature coefficients as $\bm{\gamma} = \widehat{\ma{A}}\ma{X}\tr \ma{QD}^{-1/2}\bm{\delta}$.
    \item If there are $n_{\text{test}}$ hold-out test data 
    %with the $n_{\text{test}} \times n$ kernel matrix $\ma{K}_{\text{test}}$, then $\ma{Z}_{\text{test}} = \ma{K}_{\text{test}}\ma{QD}^{-1/2}$ and 
    the linear combination for the test data becomes $\bm{\eta}_{\text{appr.test}} = \ma{X}_{\text{test}} \bm{\gamma}$. 
\end{enumerate}

In the case that one wants no penalty on the intercept by applying centring to $\ma{K}$, $\ma{X}$, and thus to $\bm{\eta}$ as in Section~\ref{sec:intercept}, one can simply read the formulas above in the computational steps with $\ma{K}$, $\ma{X}$, and thus to $\bm{\eta}$ being substituted by $\ma{K}_{\text{c}}$, $\ma{X}_{\text{c}}$, and $\bm{\eta}_{\text{c}}$ respectively. In case an inverse is needed, we replace the inverse by the Moore-Penrose inverse, for example, if $\ma{K}_{\text{c}}^{-1}$ is needed, we use $\ma{K}_{\text{c}}^{-1} = \ma{K}_{\text{c}}^- = \ma{QD}^-\ma{Q}\tr$ where $\ma{K}_{\text{c}} = \ma{QD}\ma{Q}\tr$ is the eigen decomposition of $\ma{K}_{\text{c}}$, $\ma{D}$ is the diagonal matrix with nonnegative eigenvalues, and $\ma{D}^-$ is the diagonal matrix with diagonal elements $d_{ii}^{-1}$ if $d_{ii} > 0$ and 0 otherwise.

\section{Applications}
\label{sect:application}
\vspace{-0.5cm}

\ \subsection{Chemometric data set `apples'}
\label{subsec:Apples)}
 To illustrate the use of approximated kernels, we use a data set by \cite{zude2006apples} on a chemometric study on a sample of apples. 
The explanatory variables are spectroscopic data on 256 wavelengths, denoted here simply as 1 to 256, and the response variable is soluble solids content (SSC) in units of Brix (symbol $^\circ$Bx, a measure of the dissolved solids in a liquid, representing the strength of the solution as percentage by mass).
The original data set has 642 apples, but we used a reduced sample of size $n=179$.
This ensures that there are many more variables than samples in the training set, in order to illustrate the benefit of the approximate RBF approach. 
We also use the compositional data analysis (CoDA) approach \citep{Greenacre:18, Greenacre:21} since the relative values of the chemometric data are regarded as relevant, not their absolute values.
The additive logratio transformation is applied to the data \citep{Greenacre:21b}, thereby reducing the number of predictor variables from 256 to 255. 
In a separate exercise, it was confirmed that the logratio-transformed variables perform better than the original spectroscopic variables in predicting Brix.

Having established the predictor set, we compare two methods: (a) kernel ridge regression with the radial basis function kernel (RBF) using scaling parameter $1/p$, and (b) approximated kernel ridge regression using again the RBF kernel (Approx RBF). All computations were done in \textsf{R}.
% 2025--08-07 PG: We reprogrammed it ourselves, so here we did not use glmnet.
%, {\color{blue} and methods (b) and (c) used the \texttt{glmnet}-package}. 
The optimal $\lambda$ for each of these shrinkage methods was determined by 10-fold cross validation. The same folds were used over the two methods. For each method, $100$ replications were done to see how stable the methods were over many random samples. The performance measure was the root mean squared prediction error (RMSE) 
and overall performance was summarized by boxplots of the 100 replications (Fig.~\ref{fig:RMSEApple}) for the test set results, and a table of medians over the 100 replications, for both training and test sets (Table~\ref{tab:RMSEAppleTable}). Additionally in Table~\ref{tab:RMSEAppleTable}, for each replication, the RMSE of the two methods was ranked as lowest RMSE (1) or highest (2), and their average ranks over the replications was taken, the average being preferred here because the individual values are integers. 

The kernel ridge regression performs the best, as expected, with mean rank on the test set equal to 1.38, but does not produce any estimates of effect sizes. 
%The approximate kernel ridge regression performs only marginally better than the multiple linear regression. 
The median coefficient estimates in Fig.~\ref{fig:BetasApples} show more variation in the middle spectral bands than at the extremes, showing some clear positive and negative peaks in the middle bands that predict the response variable Brix. 
%The two results in Fig.~\ref{fig:BetasApples} are very similar, with the approximated kernel ridge regression estimates shrunken on average by about 25\% compared to the multiple linear regression estimates.
Notice that the RMSEs for the RBF and the Approx RBF are identical on the training set, because the number of variables exceeds the number of samples.

\begin{knitrout}
\definecolor{shadecolor}{rgb}{0.969, 0.969, 0.969}\color{fgcolor}\begin{figure}

{\centering \includegraphics[width=10cm]{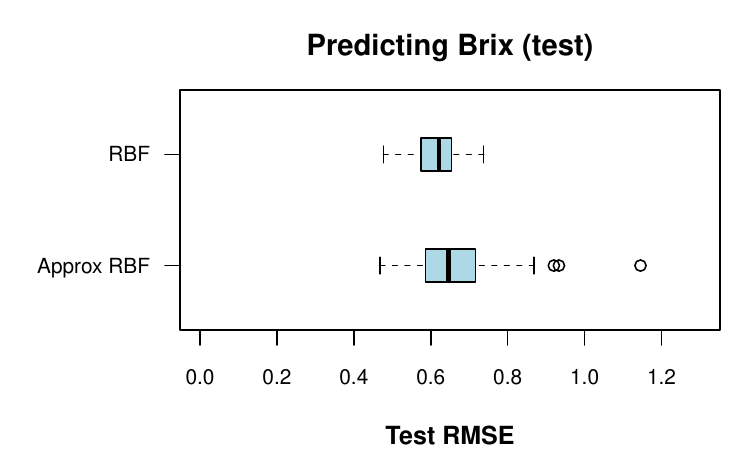} 

}

\caption[Boxplots of test RMSEs  over 100 random splits of the data in 2/3 training data and 1/3 test data]{Boxplots of test RMSEs  over 100 random splits of the data in 2/3 training data and 1/3 test data.}\label{fig:RMSEApple}
\end{figure}

\end{knitrout}

\begin{knitrout}
\definecolor{shadecolor}{rgb}{0.969, 0.969, 0.969}\color{fgcolor}\begin{table}

\caption{\label{tab:RMSEAppleTable}Median RMSE and mean ranks for the training and test sets in the prediction Brix over 100 random samples of $n = 119$ from the total sample $N  = 179$ and the remaining $n_\text{test} = 60$ as test samples, for KRR with the RBF kernel and the approximated RBF kernel. Lower values are better. }
\centering
\begin{tabular}[t]{lrrrr}
\toprule
\multicolumn{1}{c}{ } & \multicolumn{2}{c}{Median RMSE} & \multicolumn{2}{c}{Mean rank} \\
\cmidrule(l{3pt}r{3pt}){2-3} \cmidrule(l{3pt}r{3pt}){4-5}
  & Training & Test & Training & Test\\
\midrule
RBF & 0.323 & 0.621 & 1.54 & 1.38\\
Approx RBF & 0.323 & 0.646 & 1.46 & 1.62\\
\bottomrule
\end{tabular}
\end{table}

\end{knitrout}

\begin{knitrout}
\definecolor{shadecolor}{rgb}{0.969, 0.969, 0.969}\color{fgcolor}\begin{figure}

{\centering \includegraphics[width=8cm]{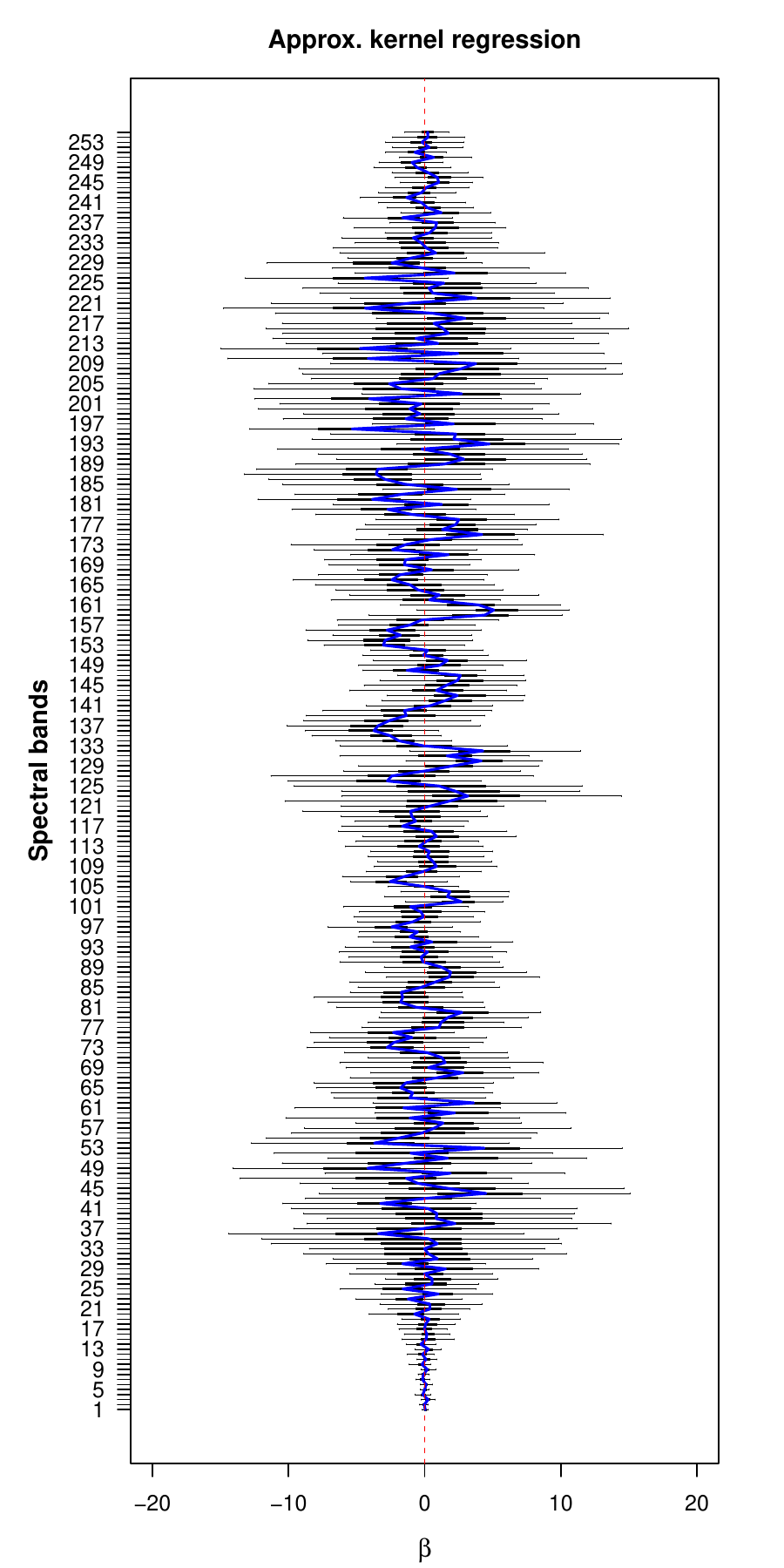} 

}

\caption[Boxplots of coefficients for regression and approximated kernel regression, to predict Brix]{Boxplots of coefficients for regression and approximated kernel regression, to predict Brix. Boxplots are obtained by randomly splitting the data a 100 times in 2/3 training data and 1/3 test data. The blue lines connect the medians of the coefficients over the replications.}\label{fig:BetasApples}
\end{figure}

\end{knitrout}

\subsection{Microbiome data set `Crohn'}
\label{subsec:Crohn}
Data set `Crohn' is available in the \textsf{R} package
\texttt{coda4microbiome} \citep{calle2023} in a modified form, with zero counts replaced (see \cite{greenacre2024chiPower}).
The data are published in their original form with zeros in the selbal R package and analysed by \cite{rivera2018} -- this original data set is analysed here. 
The data form a matrix of counts of
bacterial species aggregated into $p = 48$ genera on $n = 975$ human samples. 
In addition, each sample has been classified as having the digestive ailment called
Crohn’s disease (662 samples) or not (313 samples). 

These data are compositional, and their total count in each sample is irrelevant, and are thus expressed as proportions of their respective sample totals: $y_{ij} = x_{ij} / \sum_{j=1}^p x_{ij}$.
Furthermore, \cite{greenacre2024chiPower} has shown that a power transformation of these compositional data, including the zeros, with a power of $0.28$ (i.e., practically a fourth-root), is optimal for predicting the disease: $z_{ij} = y_{ij}^{0.28}$.
These transformed values $z_{ij}$ are used as the features in the following.
In this case where $p<n$, we include logistic regression as a third method for comparison with RBF and Approx RBF.

We did 100 random splits of the data in 2/3 training data and 1/3 test data, again stratified to respect the proportions of Crohn and non-Crohn in the whole data set. Figure~\ref{fig:AccuCrohn} shows the misclassification rate (i.e., proportion of incorrect predictions) in the test set for each of the three methods. On the test data, we see that the RBF kernel still performs best, then the approximated kernel approach followed by regular logistic regression. Ranking per replication the best performing method on the test data by 1 and the worst by 3, the average rank on the test data is 1.00 for the RBF kernel (i.e., it is the best on all replications), 2.38 for the approximated kernel approach, and 2.62 for logistic regression (Table~\ref{tab:AccuCrohnTable}).

For the Crohn data, the median KAF over the 100 replications was 
0.749,
indicating that the approximated RBF kernel accounts for approximately three quarters of the RBF kernel penalty.

\begin{knitrout}
\definecolor{shadecolor}{rgb}{0.969, 0.969, 0.969}\color{fgcolor}\begin{figure}

{\centering \includegraphics[width=10cm]{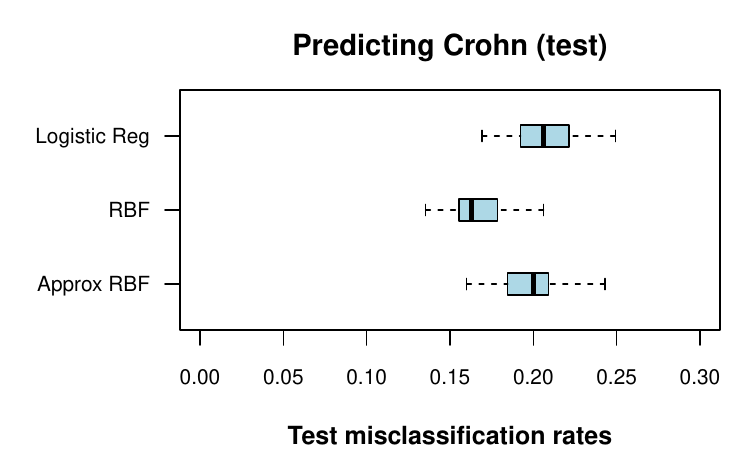} 

}

\caption[Boxplots of test misclassification rates over 100 random splits of the data in 2/3 training data and 1/3 test data]{Boxplots of test misclassification rates over 100 random splits of the data in 2/3 training data and 1/3 test data.}\label{fig:AccuCrohn}
\end{figure}

\end{knitrout}

\begin{knitrout}
\definecolor{shadecolor}{rgb}{0.969, 0.969, 0.969}\color{fgcolor}\begin{table}

\caption{\label{tab:AccuCrohnTable}Median misclassification rates and mean ranks for the training and test sets in the prediction of Crohn's disease over 100 random samples of $n = 650$ from the total sample $N  = 975$ and the remaining $n_\text{test} = 325$ as test samples, for the three methods. Lower values are better. }
\centering
\begin{tabular}[t]{lrrrr}
\toprule
\multicolumn{1}{c}{ } & \multicolumn{2}{c}{Median misclass} & \multicolumn{2}{c}{Mean rank} \\
\cmidrule(l{3pt}r{3pt}){2-3} \cmidrule(l{3pt}r{3pt}){4-5}
  & Training & Test & Training & Test\\
\midrule
Logistic Reg & 0.169 & 0.206 & 2.25 & 2.62\\
RBF & 0.029 & 0.163 & 1.00 & 1.00\\
Approx RBF & 0.173 & 0.200 & 2.75 & 2.38\\
\bottomrule
\end{tabular}
\end{table}

\end{knitrout}

\begin{knitrout}
\definecolor{shadecolor}{rgb}{0.969, 0.969, 0.969}\color{fgcolor}\begin{figure}

{\centering \includegraphics[width=\maxwidth]{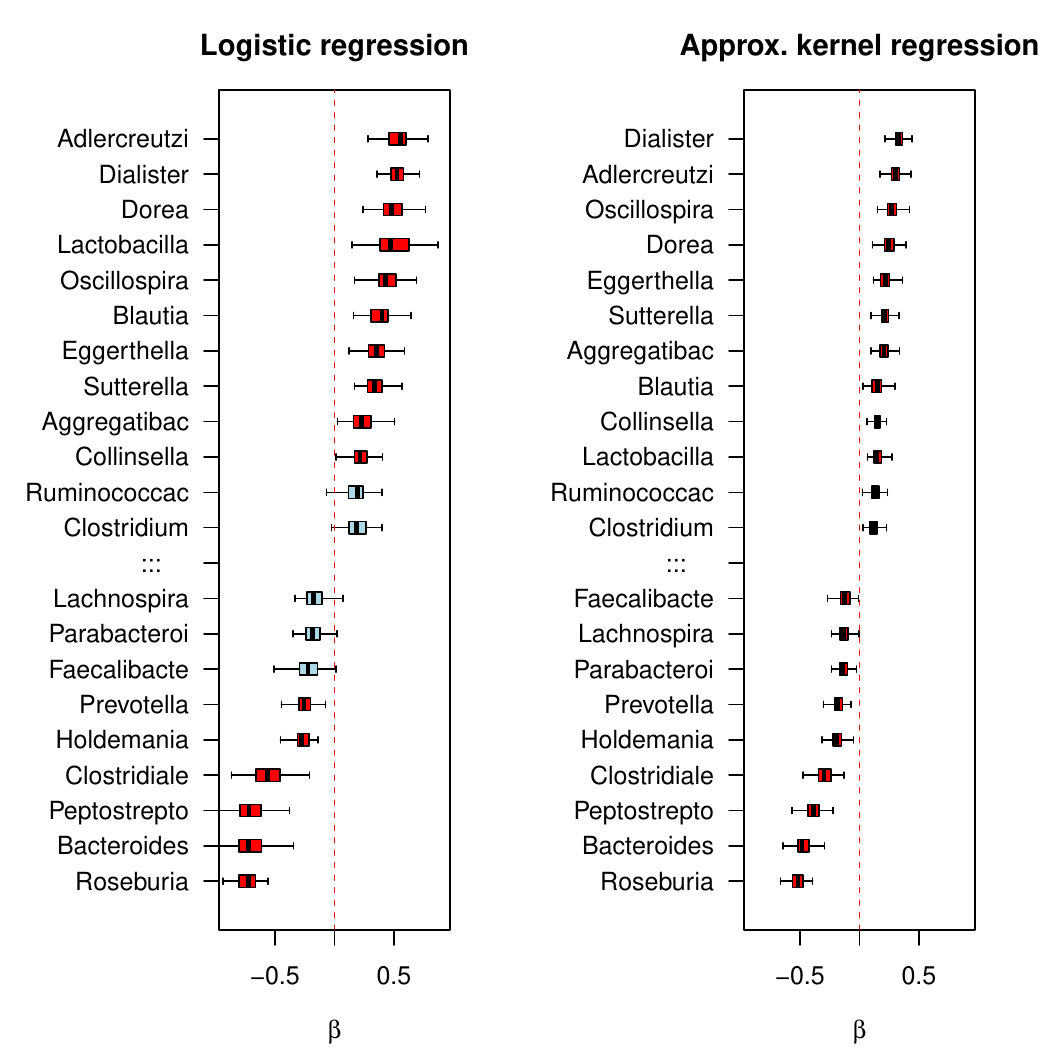} 

}

\caption[Boxplots of coefficients for logistic regression  and approximated kernel logistic regression, to predict presence or absence of Crohn's disease]{Boxplots of coefficients for logistic regression  and approximated kernel logistic regression, to predict presence or absence of Crohn's disease. Boxplots are obtained by randomly splitting the data a 100 times in 2/3 training data and 1/3 test data. Red boxes indicate that zero is not in the range of the replicate values. The figure only reports variables (bacteria) that have a red box for the approximated kernel. Some bacteria names have been shortened.}\label{fig:BetasCrohn}
\end{figure}

\end{knitrout}

The boxplots of Figure~\ref{fig:BetasCrohn} summarize the estimated coefficients for the 100 replications, for the two methods where coefficients can be computed.
It can be seen in both methods that the increased presence of bacteria Dialister and several others consistently predict higher probability of Crohn's disease, whereas increased presence of bacteria such as Roseburia and Bacteroides consistently predict less chance of the disease. 
The estimated coefficients for the approximated kernel logistic regression are a shrunken version of those for the logistic regression (Fig.~\ref{fig:BetasCrohn}).
Moreover, the dispersions of the estimates over replications
in the right hand plot in Fig.~\ref{fig:BetasCrohn} are seen to be much lower.

Finally, as a comparison, \cite{greenacre2024chiPower} performed variable selection on the same data set, using the same power transformation in the logistic regression, and chose only 14 out of the 48 predictors as being significant. 
There the misclassification rates of 0.184 (training set) and 0.214 (test set) were found, using only one replication of the same random training/test set split.
These rates are within the bounds of the replications of Fig.~\ref{fig:AccuCrohn}, but both slightly worse than the median values of 0.169 and 0.206 in Table \ref{tab:AccuCrohnTable} using all 48 predictors.  
The 14 selected predictors coincide exactly with the 8 top positive predictors and 6 top negative predictors in the boxplots in the logistic regression results of Fig.~\ref{fig:BetasCrohn}, all of which have dispersions well separated from the zero line.
%Notice that the predictors have been ordered according to the results of the approximated kernel regression on the right, and that \textsf{Lactobacilla} is included in the top 8 positive predictors in the logistic regression on the left.

\section{Discussion and conclusion}
\label{sect:conclusions}
The present paper shows through linear algebra that it is possible with a wide predictor matrix to reconstruct exactly predictions obtained by kernel with a ridge penalty, and approximately so if the predictor matrix is tall.

%in the context of have a similar objective to ours, deriving what they call effect size analogs using an approximated kernel matrix.
A caveat is that the numeric computations depend on inverses and eigen decompositions of size $n \times n$. The order of operations needed to compute these are $O(n^3)$, so that computationally, our results get harder to compute as $n$ gets into the ten thousands. For those cases, approximate kernels could be used. Section \ref{sect:reexpressKernelAsRegularRidge} reexpressed the kernel penalty as a regular ridge penalty with predictor variables being the principal components of the kernel matrix. Therefore, efficient approximations could be done by first finding a limited number of principal components of the kernel matrix and use the methodology of this paper to ensure that the approximated kernel matrix is as close as possible to the kernel matrix. We leave this as a topic for further research.

In line with the nonparametric approaches often taken in the machine learning literature, we relied on sampling approaches to establish coefficients to be statistically different from zero. For kernel generalized linear models, it would also be possible to derive $p$-values for the coefficients based on statistical tests that make use of effective degrees of freedom. %, which we have left for future research.  
%\cite{apley2020effects} propose accumulated local dependence (ALE) plots as an alternative to the commonly used partial dependence (PD) plots...  \textbf{[Do we have to explain what these are?]}
%\cite{rudin2019blackbox} pleas for XAI when critical decisions depend on the results of a black box prediction model, and also favours the use of ALE over PD plots to visualize effects in a supervised learning context. 

%\bibliographystyle{plain} % basic style, author-year citations
\bibliographystyle{abbrvnat}
\bibliography{XAI}        % name your BibTeX data base

\end{document}